# Improving Cryptocurrency Pump-and-Dump Detection through Ensemble-Based Models and Synthetic Oversampling Techniques

Jieun Yu[1]  (Ewha Womans University)

Minjung Park[2]  (Kumoh National Institute of Technology)

Sangmi Chai[3]  (Ewha Womans University)

Abstract

This study aims to detect pump-and-dump (P&D) manipulation in cryptocurrency markets, where the scarcity of such events causes severe class imbalance and hinders accurate detection. To address this issue, the Synthetic Minority Oversampling Technique (SMOTE) was applied, and advanced ensemble learning models were evaluated to distinguish manipulative trading behavior from normal market activity. The experimental results show that applying SMOTE greatly enhanced the ability of all models to detect P&D events by increasing recall and improving the overall balance between precision and recall. In particular, XGBoost and LightGBM achieved high recall rates (94.87% and 93.59%, respectively) with strong F1-scores and demonstrated fast computational performance, making them suitable for near real-time surveillance. These findings indicate that integrating data balancing techniques with ensemble methods significantly improves the early detection of manipulative activities, contributing to a fairer, more transparent, and more stable cryptocurrency market.

[1]  kochstje@naver.com
[2]  mjpark@kumoh.ac.kr
[3]  smchai@ewha.ac.kr, corresponding author

1. Research Introduction

The emergence of cryptocurrency markets has introduced new forms of financial innovation while simultaneously exposing investors to unprecedented risks arising from the absence of centralized regulation and the rapid diffusion of speculative assets. Among the various types of illicit activities in this domain, pump-and-dump (P&D) schemes have become a significant concern due to their highly organized nature and their disproportionate impact on retail investors. In these schemes, coordinated groups artificially inflate the value of a low-liquidity digital asset through concentrated buying and aggressive dissemination of promotional messages across social media and messaging platforms such as Telegram and Discord (Hamrick et al., 2018). Once the price reaches an artificially created peak, these groups liquidate their holdings, causing the price to collapse and leaving late-arriving investors with substantial losses. Such manipulative practices not only undermine market integrity but also erode public trust in emerging digital asset ecosystems, ultimately hindering the sustainable growth of financial innovation.

The social consequences of P&D manipulation have been well-documented. A comprehensive analysis by The Wall Street Journal reported that between January and June 2018, 175 coordinated P&D events across 121 cryptocurrencies generated an estimated USD 825 million in trading volume, inflicting hundreds of millions of dollars in losses on retail investors. The U.S. Commodity Futures Trading Commission (CFTC) initiated its first enforcement action against digital-asset pump-and-dump manipulation in 2021, while the U.S. Securities and Exchange Commission (SEC) has also charged firms for large-scale manipulative schemes that misled investors into committing tens of millions of dollars (CFTC, 2021; SEC, 2022). These high-profile interventions highlight the systemic vulnerabilities of digital asset markets, where the combination of anonymity, fragmented global trading venues, and limited oversight creates fertile ground for organized manipulation.

Despite the growing attention of regulators, academic research on the detection of cryptocurrency market manipulation remains at an early stage. Existing studies, such as Hamrick et al. (2018) and La Morgia et al. (2021), have provided valuable empirical insights into the prevalence and structural characteristics of P&D schemes. However, most of these studies have been retrospective in nature, relying on descriptive statistics or rule-based heuristics to document manipulative behavior after it occurs. This approach, while useful for forensic analysis, is insufficient to enable timely detection and preventive action. Furthermore, P&D events account for only a very small proportion of market activity, producing highly imbalanced datasets that degrade the performance of conventional detection algorithms. Without addressing this imbalance, classification models tend to be biased toward the majority (non-manipulated) class, making them unreliable as real-time surveillance tools.

To address these shortcomings, this study focuses on developing machine learning-based detection models that are specifically designed to identify pump-and-dump activities in cryptocurrency markets. We incorporate the Synthetic Minority Oversampling Technique (SMOTE) to correct severe class imbalance and systematically evaluate multiple ensemble learning algorithms to determine their effectiveness in distinguishing manipulative events from normal market activity. By integrating data-level resampling and model-level ensemble techniques, the study aims to contribute a methodology that is both robust and practically applicable in the context of decision support systems for exchanges and regulatory authorities. In doing so, it seeks to support fairer, more transparent digital asset markets and to reduce the systemic risks posed by manipulative trading practices. This research therefore responds to a pressing societal need, bridging the gap between post-hoc analysis and proactive detection, and represents a significant step toward operational tools that can protect market participants in the rapidly evolving cryptocurrency ecosystem.

2. Related Works

2.1 Market Manipulation in Cryptocurrency Markets

Research on market manipulation in cryptocurrency markets has expanded significantly over the past decade, reflecting the rapid evolution and global expansion of digital asset trading. Unlike regulated equity markets, cryptocurrency markets operate in a fragmented and lightly supervised environment, which fosters conditions conducive to manipulation. Structural characteristics—decentralized and cross-border exchanges, high volatility, and the absence of centralized clearing—create an ecosystem in which the actions of a few coordinated actors can distort asset prices, mislead market participants, and erode overall trust in the market. These features have made cryptocurrency market manipulation an urgent focus for academics, regulators, and practitioners alike.

Early studies relied on exchange-level transaction data to investigate suspicious trading behavior. For example, Gandal et al. (2018) analyzed Mt. Gox records and showed that automated trading bots, apparently operated by insiders, had a significant and positive correlation with sharp Bitcoin price surges, suggesting that manipulative trading practices could strongly affect market prices even without formal market-making infrastructure. Similarly, Kong & Schoenebeck (2018) experimentally demonstrated that frequent, low-value micro-transactions executed by bots triggered imitative buying behavior from other market participants, thereby amplifying short-term volatility. Griffin and Shams (2020) further examined the relationship between the issuance of Tether, the largest USD-pegged stablecoin, and Bitcoin price movements, providing evidence that sudden increases in Tether circulation are closely linked to subsequent surges in cryptocurrency prices. This line of research underscored the systemic risks introduced by stablecoins and their potential use as a vehicle for coordinated price inflation.

Beyond centralized exchanges, manipulation has expanded to decentralized finance (DeFi) and privacy-enhancing services, broadening the spectrum of abusive strategies. Mixing services and privacy wallets such as Wasabi Wallet and Tornado Cash have been shown to facilitate obfuscation of transaction trails, enabling manipulators to conceal the origin of funds and to coordinate wash trades and pump-and-dump activities with reduced traceability (Mariani & Homoliak, 2025). Such mixers are increasingly used in conjunction with decentralized exchanges, allowing actors to bypass surveillance mechanisms that were previously more effective on centralized platforms.

Another form of manipulation involves fraudulent token schemes such as scams, rug pulls, and misleading initial coin offerings (ICOs). Researchers have noted that new or obscure tokens with very low liquidity are particularly prone to scam-driven manipulation, where developers create hype, inflate the price through misleading promotions, and then abandon the project by removing liquidity—commonly referred to as "rug pulls" (Pradeep et al., 2022; Zhou et al., 2024). Unlike classic P&D events, these schemes typically operate on decentralized platforms, exploit the absence of counterparty due diligence, and leave retail investors with irrecoverable losses. This variant of manipulation represents a structural shift from centralized-market behavior to on-chain programmable fraud.

The classic pump-and-dump (P&D) model remains a dominant form of market manipulation, but it has evolved through the use of modern communication channels. Hamrick et al. (2018) compiled one of the first large-scale datasets documenting P&D events organized through Telegram and Discord channels. They showed that such events produce extreme, short-lived price spikes and trading volumes, followed by rapid price collapses as organizers liquidate their holdings. La Morgia et al. (2021) extended this line of inquiry by monitoring these communities in real time, documenting the typical lifecycle of P&D manipulation and identifying behavioral markers such as synchronized large market orders ("rush orders") and coordinated social media campaigns. These behaviors have been observed not only on centralized exchanges but also in decentralized ecosystems, with on-chain messaging and automated bots amplifying their speed and impact.

Together, these studies reveal that cryptocurrency markets are exposed to a broad spectrum of manipulative strategies, including wash trading, spoofing, insider trading, pump-and-dump campaigns, liquidity manipulation through stablecoins, scam tokens, and obfuscation through coin mixers and privacy wallets. While these manipulations differ in their mechanics, they share common characteristics: they exploit the lack of regulatory oversight, capitalize on asymmetric information, and rely on the high speed and anonymity afforded by blockchain systems.

Recent studies collectively point to several important implications. First, it shows that the scope of cryptocurrency manipulation has expanded from centralized exchange-based behaviors to include decentralized systems, making detection increasingly complex. Second, the speed and sophistication of these manipulative activities emphasize the need for proactive, real-time monitoring tools and data-driven detection models that can work across diverse trading environments. While most existing studies have focused on post-event analysis and retrospective descriptions of manipulation, this study aims to move beyond that limitation by developing predictive approaches designed to anticipate and detect manipulative events at an earlier stage, thereby reducing their potential impact on retail investors and overall market stability.

## 2.2 Pump-and-Dump

Among the various forms of market manipulation identified in cryptocurrency trading, pump-and-dump (P&D) schemes represent one of the most pervasive and damaging behaviors. Unlike other manipulative practices such as insider trading, spoofing, or wash trading, which often require sophisticated infrastructure or privileged access to market information, P&D events rely primarily on rapid coordination among groups of actors who use low-liquidity assets as targets. These schemes typically begin with the organizers disseminating signals to private communication channels such as Telegram and Discord, encouraging members to place large market buy orders simultaneously. This sudden, coordinated buying activity creates an artificial surge in price and trading volume. Once external investors—driven by fear of missing out (FOMO)—enter the market, the organizers liquidate their holdings at inflated prices, triggering a rapid decline and leaving late entrants with substantial losses (Balcilar & Ozdemir, 2023).

Empirical studies have described the mechanisms and lifecycle of P&D events in both centralized and decentralized trading environments. La Morgia et al. (2021) further examined the organizational dynamics of these communities, documenting that early participants, who receive instructions in advance, benefit disproportionately, while most retail investors suffer losses due to delayed participation. These studies highlight how P&D schemes exploit behavioral biases such as herd mentality and FOMO, which are amplified in fast-moving, lightly regulated markets.

Although other manipulative behaviors such as wash trading or rug pulls can cause systemic harm, this study focuses on pump-and-dump events for several reasons. First, P&D schemes have a high frequency and broad market impact: compared to insider trading or rug pulls, which are less frequent but often larger in scale, P&D events occur daily and can simultaneously affect dozens of small-cap assets, causing volatility spikes that spread across the market (Fantazzini & Xiao, 2023). Second, P&D manipulation leaves a distinct, time-compressed signature in price and volume patterns, making it an ideal candidate for machine learning models that analyze high-frequency data. The availability of historical P&D datasets collected from community channels (e.g., Hamrick et al., 2018; La Morgia et al., 2021) enables systematic model training and validation, which is often not possible for rarer forms of manipulation such as rug pulls. Third, P&D events unfold extremely quickly, often in less than 15 minutes, creating an urgent need for detection models that can provide real-time alerts to mitigate harm. Unlike other frauds, which may be uncovered over weeks or months, the prevention of damage from P&D requires models that can operate on sub-hourly data streams.

For these reasons, this study focuses on developing machine learning models for the detection of P&D manipulation. In particular, it addresses the severe class imbalance inherent in transaction-level data through oversampling techniques, with the goal of constructing models capable of more accurately identifying manipulative behaviors. Through this approach, the study aims to develop detection models that can serve as a core component of real-time surveillance systems in digital asset markets.

2.3 Machine Learning Approaches for Anomaly and Fraud Detection

Machine learning (ML) and artificial intelligence (AI) techniques have long been central to the detection of fraudulent and anomalous patterns in financial transactions. In domains such as credit card fraud detection, insider trading detection, and anti-money-laundering surveillance, the complex, high-dimensional nature of transactional data makes traditional rule-based systems insufficient. ML methods, by contrast, can model non-linear patterns and adapt to changing behaviors, which has led to their broad application in financial anomaly detection.

Early surveys by Phua et al. (2010) and Ngai et al. (2011) classified machine learning–based fraud detection systems into supervised, semi-supervised, and unsupervised paradigms, demonstrating that data-driven approaches often outperform static rule sets. In supervised learning, labeled datasets of fraudulent and legitimate activities enable classifiers such as decision trees, support vector machines, and ensemble algorithms to learn distinguishing characteristics. In semi-supervised approaches, models are trained predominantly on legitimate transactions to learn a baseline of normal behavior, after which deviations are flagged as anomalies. Unsupervised learning, on the other hand, is particularly valuable when labeled data are scarce; clustering and density-based methods can detect outliers in transaction patterns without prior labeling.

Among these methods, ensemble techniques have proven especially robust in financial contexts. Random forests aggregate multiple decision trees to reduce overfitting and improve generalization. Boosting methods, including AdaBoost, Gradient Boosting Machines (GBM), XGBoost, and LightGBM, iteratively focus on misclassified cases, achieving high predictive performance in imbalanced and noisy datasets. These models excel at capturing non-linear relationships and interactions, making them particularly suited for domains where the fraudulent patterns evolve quickly.

Recent applications of machine learning to financial markets include detecting insider trading through unusual trading patterns, classifying credit card fraud in real time, and identifying anomalous orders in high-frequency trading systems. These studies demonstrate the versatility of ensemble methods in improving early warning systems. However, the application of such approaches to cryptocurrency market manipulation remains relatively limited. While Xu and Livshits (2019) and La Morgia et al. (2021) proposed models to predict P&D schemes, these efforts often relied on raw price and volume metrics as features and did not leverage the full potential of advanced learning methods.

Moreover, traditional machine learning research in finance typically assumes stable market structures and availability of reliable labels, conditions that do not hold in decentralized cryptocurrency markets. The high volatility, fragmented trading venues, and rapidly changing behavioral patterns of manipulators require models capable of adapting to complex and non-stationary environments. This research therefore aims to extend machine learning methodologies developed in conventional financial markets into the unique and challenging setting of cryptocurrency markets, where the dynamics of manipulation differ significantly from those observed in regulated equities.

3. Research Methodology

3.1 Ensemble Learning

Ensemble learning refers to a family of techniques that combine multiple base models to improve predictive accuracy and robustness in decision-making tasks. Rather than relying on a single model, ensembles exploit the diversity among individual models so that errors made by one can be compensated for by others. This approach is especially relevant in complex domains such as financial anomaly detection, where patterns are nonlinear and data distributions can be unstable. Bagging and boosting are two dominant strategies: bagging constructs models in parallel on bootstrapped samples to reduce variance, while boosting builds models sequentially, focusing on correcting errors from previous iterations to reduce bias (Dietterich, 2002; Polikar, 2006). Tree-based ensembles, such as Random Forests and Gradient Boosting Machines, have been shown to balance interpretability and predictive power, making them ideal for applications where both explainability and generalization are crucial (Sagi & Rokach, 2018; Ganaie & Hu, 2021). In the context of cryptocurrency pump-and-dump detection, ensembles provide resilience against noisy signals caused by market volatility.

In this study, ensemble-based models were exclusively selected because prior research has consistently shown that ensemble methods outperform single learners in imbalanced and noisy financial datasets, particularly when identifying rare, high-impact events (Zhou, 2012; Liu et al., 2008). Furthermore, deep learning models were not adopted due to the limited size of labeled data available for pump-and-dump events, the risk of overfitting in highly imbalanced settings, and the need for faster training and inference times for real-time detection. Given these considerations, ensemble models offer an optimal balance of accuracy, efficiency, and interpretability, making them well-suited for the early detection of pump-and-dump manipulation in cryptocurrency markets.

### 3.2 Random Forest

Random Forest, a bagging-based ensemble, generates multiple decision trees by randomly sampling both the data and the predictor variables (Breiman, 2001). By incorporating randomness in feature selection at each split, Random Forest reduces the correlation between trees and lowers overfitting risk. The final prediction is aggregated through majority voting or averaging, depending on whether the task is classification or regression. This method is robust to noise and can handle datasets with a large number of features, which is particularly important when working with transaction data characterized by diverse attributes such as trade volume, price volatility, and time-series patterns. The technique also provides feature importance measures, which facilitate the interpretation of which variables drive model predictions, though the overall model remains more complex than single decision trees.

### 3.3 AdaBoost (Adaptive Boosting)

AdaBoost transforms weak learners—models that perform only slightly better than random guessing—into a strong classifier by training them sequentially, with a focus on misclassified observations in each subsequent iteration (Freund & Schapire, 1997). Each weak learner is assigned a weight based on its classification error, and the ensemble prediction is computed as a weighted sum of all weak learners. While simple and computationally efficient, AdaBoost is sensitive to noise and outliers because misclassified samples receive increasing emphasis. For financial anomaly detection, this sensitivity can be both a strength and a limitation: it highlights subtle patterns in minority classes but risks overfitting when the data contains spurious fluctuations.

### 3.4 Gradient Boosting Machine (GBM)

Gradient Boosting Machines improve on the general boosting framework by using gradient descent to minimize a differentiable loss function (Friedman, 2001). Instead of adjusting sample weights like AdaBoost, GBM trains each new model to predict the residual errors of the previous ensemble, iteratively reducing bias. This method is flexible because it allows the use of different loss functions, making it adaptable to classification or regression problems. However, GBM models can be slow to train and prone to overfitting without proper regularization. Their strength lies in handling complex non-linear relationships, which is critical in detecting manipulative trading patterns where subtle market signals may indicate pump-and-dump behavior.

### 3.5 Extreme Gradient Boosting (XGBoost)

XGBoost is an optimized version of GBM designed to enhance computational efficiency and regularization. By using second-order gradient information and supporting parallel computation, XGBoost achieves faster convergence while controlling overfitting (Chen & Guestrin, 2016). Additional techniques, such as handling missing values internally and using early stopping, make it practical for large-scale and high-dimensional datasets. XGBoost has become widely adopted in financial machine learning tasks, as it balances speed and accuracy, which is crucial in environments like cryptocurrency markets where rapid model retraining and deployment may be required.

### 3.6 Light Gradient Boosting Machine (LightGBM)

LightGBM, like XGBoost, is derived from GBM but introduces histogram-based decision tree construction and a leaf-wise growth strategy with depth constraints (Ke et al., 2017). These optimizations significantly reduce memory usage and training time, making it ideal for massive datasets with high feature dimensionality. By focusing on leaves with the largest gradient changes, LightGBM captures complex interactions among features more effectively, which is advantageous for modeling the high-frequency, nonlinear nature of cryptocurrency trading data.

### 3.7 Oversampling for Class Imbalance

One of the most significant challenges in pump-and-dump detection is the extreme class imbalance: manipulative events occur infrequently relative to normal trading activity. Models trained on imbalanced data tend to be biased toward the majority class, resulting in poor recall for rare but critical events. To address this, the Synthetic Minority Oversampling Technique (SMOTE) was employed in this study (Chawla et al., 2002). SMOTE generates synthetic minority samples by interpolating between existing minority instances, thereby creating a more balanced training set. This approach has been widely validated in fraud detection, medical diagnosis, and credit risk assessment and has shown to significantly enhance model sensitivity to minority events. In the context of this research, applying SMOTE prior to training ensures that the ensemble models can focus on learning distinguishing characteristics of pump-and-dump activity, improving recall without overly compromising precision.

## 4. Experimental Design

### 4.1 Objectives and Overview of Experiment

This study conducts a series of experiments to construct a detection model capable of identifying pump and dump (P&D) events that occur frequently in cryptocurrency exchanges. A central challenge of this dataset is its extreme class imbalance: legitimate trading activity overwhelmingly dominates the data, while the manipulative events of interest constitute only a very small fraction of the observations. To address this imbalance, the Synthetic Minority Oversampling Technique (SMOTE) was adopted as part of the data preprocessing pipeline. Five tree based ensemble algorithms were selected as candidate models for experimentation: Random Forest, AdaBoost, Gradient Boosting Machine (GBM), XGBoost, and LightGBM. Two sets of experiments were conducted: 1) Models trained on the original imbalanced dataset, and 2) Models trained on the SMOTE augmented dataset. The results of these experiments were compared to evaluate the impact of oversampling on model performance and to assess whether oversampling contributes to a more effective detection framework for P&D events.

4.2 Experimental Data

For the empirical analysis, this study employs the open dataset published by La Morgia et al. (2020). This dataset contains 317 documented pump-and-dump (P&D) events collected from the Binance exchange and covers a total of 85 cryptocurrency tokens. La Morgia et al. (2020) actively participated in 19 P&D groups on Telegram and Discord to collect event schedules and used the official Binance API to extract detailed trading records corresponding to those manipulation windows. Binance was chosen as the data source because of its large active user base and high trading volume, which makes it one of the most liquid and dynamic markets; these characteristics present a challenging yet realistic environment for detecting manipulative activities.

In their original work, La Morgia et al. (2020) constructed three datasets by varying the chunk size and time window used for feature extraction. Their primary objective was to enable early detection of P&D events; thus, they experimented with chunk sizes of 5 seconds, 15 seconds, and 25 seconds, demonstrating that shorter intervals capture manipulation dynamics more effectively. Following their results, the present study adopts the dataset based on 25-second chunks, which was found to achieve the highest predictive performance. This study differentiates itself from previous research in two key ways. First, it incorporates SMOTE oversampling into the experimental design. By applying SMOTE to the training data, the study compares models trained with and without oversampling to examine how this approach mitigates extreme class imbalance and affects predictive performance. Through this comparison, the research demonstrates the value of oversampling techniques, particularly in cryptocurrency market datasets where manipulative events occur infrequently. Second, the study broadens the range of ensemble classifiers considered. Whereas prior work focused mainly on Random Forest and logistic regression, this research extends the scope to include a more diverse set of advanced ensemble methods. Specifically, Random Forest, AdaBoost, Gradient Boosting Machine (GBM), XGBoost, and LightGBM are developed and evaluated to provide a comprehensive understanding of model performance. For each P&D event, trading data were extracted over a 14-day period (±7 days around the event date). The raw transaction logs, which include UNIX timestamps, trading volume, transaction price, and order type, were aggregated into 25-second chunks. Subsequently, features were generated using a 7-hour sliding time window to capture temporal dependencies.

The final dataset consists of 482,157 records, of which 481,840 correspond to normal trading and 317 correspond to P&D events, resulting in an imbalance ratio well over 9:1. This severe imbalance underscores the need for resampling methods. Therefore, SMOTE oversampling was applied exclusively to the training dataset, ensuring that the test dataset remained unbiased for model evaluation.

| Category | General | Pump & Dump | Total |
|---|---|---|---|
| Frequency | 481,840 | 317 | 482,157 |
| Percentage (%) | 99.93 | 0.07 | 100 |

Table 1 Distribution of General and Pump & Dump Cases

| Dataset Type | Abnormal | Normal | Total |
|---|---|---|---|
| Training | 222 | 337,287 | 144,648 (30%) |
| Test | 95 | 144,553 | 337,509 (70%) |
| Total | 317 | 481,840 | 482,157 |

Table 2 Dataset Split (Training vs. Test)

| Category | Abnormal | Normal |
|---|---|---|
| Before Oversampling | 222 | 337,287 |
| After Oversampling | 337,287 | 337,287 |
| Total | 337,509 | 674,574 |

Table 3 Data Distribution Before and After Oversampling

| Variable Name | Description |
|---|---|
| StdRushOrders | Moving standard deviation of market order trade volumes |
| AvgRushOrders | Moving average of market order trade volumes |
| StdTrades | Moving standard deviation of the number of trades |
| StdVolumes | Moving standard deviation of trade volumes |
| AvgVolumes | Moving average of trade volumes |
| StdPrice | Moving standard deviation of closing prices |
| AvgPrice | Moving average of closing prices |
| AvgPriceMax | Moving average of highest prices |
| AvgPriceMin | Moving average of lowest prices |

Table 4 Description of Variables

### 4.3 Model Evaluation Metrics

The performance of the classification models in this study was evaluated using four widely accepted metrics—Accuracy, Precision, Recall, and F1 score—calculated from the confusion matrix. These metrics provide a comprehensive assessment of model performance, which is particularly important for detecting rare events such as pump-and-dump schemes in highly imbalanced datasets.

The confusion matrix summarizes model predictions into four categories (true positives, false positives, true negatives, and false negatives) and serves as the basis for all subsequent metrics. Accuracy represents the overall proportion of correct predictions, but in imbalanced datasets it can be misleading, as a model that predominantly predicts the majority class may appear accurate. To overcome this limitation, precision and recall are also reported: precision measures the proportion of correct positive predictions, while recall measures the proportion of actual positive cases that were correctly identified. These metrics are complementary—precision reduces false alarms, and recall reduces missed detections. The F1 score, the harmonic mean of precision and recall, provides a single balanced indicator, which is especially useful when normal transactions dominate the data and the target is a small set of manipulative events. By relying on these metrics, the evaluation considers not only overall accuracy but also the model's ability to detect rare but high-impact P&D events.

### 5. Results

The experimental results reveal clear differences between models trained on the original dataset and those trained on the SMOTE-balanced dataset. In the case of the original imbalanced data, most models, with the exception of LightGBM, exhibited higher precision than recall. This pattern indicates that the models were generally more conservative, favoring correct identification of the majority class at the expense of missing a portion of the minority class events. When SMOTE oversampling was applied, however, all five models demonstrated a consistent improvement in recall, with recall scores exceeding precision in every case. This shift directly reflects the purpose of this study, which prioritizes the detection of minority-class events—pump-and-dump manipulations—where the cost of false negatives is far greater than that of false positives.

The comparison between models trained on the original and oversampled datasets highlights the value of data balancing in this domain. For example, the Random Forest model achieved the highest F1-score on the original data (93.2%), characterized by extremely high precision (98.6%) but lower recall (88.5%), illustrating the typical imbalance-driven bias toward the majority class. After applying SMOTE, the same model showed a more balanced outcome: recall increased to 93.6% and precision decreased slightly to 89.0%, demonstrating that oversampling improved the model's ability to detect rare events. A similar trend was observed for AdaBoost, GBM, and XGBoost, all of which achieved notable recall improvements ranging from 4% to 10%. Particularly noteworthy are the results from XGBoost and LightGBM, which displayed not only substantial gains in recall but also very fast training times. LightGBM, in particular, exhibited an increase of approximately 10% in recall and 7% in precision, resulting in an overall F1-score improvement of about 9%. Moreover, the computational efficiency of these two models is remarkable: XGBoost completed training in approximately 14 seconds and LightGBM in just over 3 seconds, far faster than the other models, with Random Forest requiring nearly 8 minutes and GBM almost 20 minutes.

Overall, these findings demonstrate that SMOTE oversampling significantly mitigates the challenges posed by extreme class imbalance, enabling the models to focus more effectively on detecting pump-and-dump events. While in some models the F1-score slightly decreased due to a reduction in precision, the substantial improvement in recall aligns with the primary research objective of early and reliable detection of manipulative trading behaviors. In addition, the results suggest that modern gradient boosting methods, such as XGBoost and LightGBM, offer a compelling balance between detection accuracy and computational efficiency, making them promising tools for real-time surveillance applications in cryptocurrency markets. The detailed classification performance of each model, including accuracy, precision, recall, and F1 scores, is presented in Table 5, while the corresponding training times for the SMOTE-augmented datasets are summarized in Table 6 to highlight differences in computational efficiency.

| Model | Data | Accuracy | Precision | Recall | F1-Score |
|---|---|---|---|---|---|
| RF | Original | 99.99 | 98.57 | 88.46 | 93.24 |
| RF | SMOTE | 99.99 | 89.02 | 93.59 | 91.25 |
| AdaBoost | Original | 99.99 | 92.11 | 86.42 | 89.17 |
| AdaBoost | SMOTE | 99.99 | 86.90 | 90.12 | 88.48 |
| GBM | Original | 99.99 | 92.11 | 86.42 | 89.17 |
| GBM | SMOTE | 99.98 | 82.02 | 90.12 | 85.88 |
| XGBoost | Original | 99.98 | 86.84 | 84.62 | 85.71 |
| XGBoost | SMOTE | 99.99 | 82.22 | 94.87 | 88.10 |
| LightGBM | Original | 99.98 | 77.38 | 83.33 | 80.25 |
| LightGBM | SMOTE | 99.99 | 83.91 | 93.59 | 88.48 |

Table 5 Model Performance Evaluation Metrics (Unit: %)

| Model | Training Time |
|---|---|
| RF | 474.96 (approx. 8 min) |
| AdaBoost | 31.77 |
| GBM | 1203.54 (approx. 20 min) |
| XGBoost | 14.46 |
| LightGBM | 3.33 |

Table 6 Training Time on SMOTE Data by Model (Unit: seconds)

6. Discussions

   6.1 Key Findings

This study set out to develop machine learning models capable of detecting pump-and-dump (P&D) events in cryptocurrency markets, where the prevalence of imbalanced data poses a major challenge to predictive modeling. The empirical findings demonstrate that models trained with SMOTE oversampling consistently achieved substantial improvements in recall across all ensemble methods tested. This outcome confirms the value of addressing class imbalance in scenarios where the cost of false negatives—failing to identify manipulative activity—is more severe than false positives. While precision occasionally declined after oversampling, the overall balance between precision and recall improved, resulting in models better suited to identifying rare but critical P&D events. Among the ensemble models, XGBoost and LightGBM not only provided strong predictive accuracy but also demonstrated remarkable computational efficiency, making them well suited for applications that require timely analysis, such as real-time surveillance. Furthermore, the experiments highlight that even in highly volatile, lightly regulated markets, systematic approaches based on advanced ensemble learning methods can uncover subtle patterns of manipulation that traditional monitoring systems are likely to miss. These results support the argument that advanced analytics can play a central role in strengthening the integrity and stability of cryptocurrency markets.

   6.2 Contributions

The results of this study provide several meaningful contributions that extend beyond technical modeling. The combination of SMOTE oversampling with advanced ensemble learning models demonstrates how careful preprocessing can substantially improve the sensitivity of classifiers to rare, manipulative trading patterns. By systematically comparing five ensemble techniques on event-labeled transaction data, this research shows that imbalanced-data handling is not a marginal adjustment but a critical step that directly influences the model's ability to detect market abuse in complex, noisy trading environments. These findings offer a replicable approach that other researchers and practitioners can build upon in future anomaly detection work.

      In the domain of digital asset markets, the models developed in this study present a practical pathway for identifying pump-and-dump schemes at an early stage. Reliable detection mechanisms such as these are essential for mitigating systemic risks in cryptocurrency markets, which are still largely unregulated and vulnerable to manipulation. By showing that detection performance can be significantly improved without excessive computational cost, this study highlights how similar frameworks could be embedded into surveillance infrastructures of exchanges or deployed by supervisory authorities to enhance market transparency and investor protection.

      The implications also extend to the broader context of governance and risk management. As digital assets become more integrated into corporate financing, internal control, and portfolio management, the anomaly detection framework proposed here can be adapted to monitor abnormal

transaction flows within organizations, identify irregularities in real time, and strengthen compliance structures. This integration of machine learning models into governance processes offers a way for financial institutions, auditors, and compliance teams to combine technological innovation with more robust oversight mechanisms. Ultimately, these findings show that advanced AI methodologies can contribute to shaping a safer, more trustworthy digital asset ecosystem by aligning technological precision with financial stability and responsible management practices.

### 6.3 Limitations & Future Research

While this study demonstrates the potential of ensemble learning combined with oversampling techniques for improving the detection of pump-and-dump activities, several limitations should be acknowledged. The models were developed with the primary goal of establishing an effective and reproducible framework for identifying manipulative trading behaviors in digital asset markets. However, this focus also imposes certain constraints. The dataset, though comprehensive, is restricted to a single exchange (Binance) and a specific set of P&D events. As a result, the models may reflect characteristics unique to this market and may not generalize seamlessly to other trading platforms or to different forms of misconduct such as wash trading, front-running, or rug-pull schemes. Furthermore, the analysis relies exclusively on transaction-level data; incorporating additional modalities such as social media signals, blockchain network metrics, or sentiment indicators could provide a richer context and enhance predictive capability. Finally, while the use of SMOTE helped address the severe class imbalance inherent in the data, oversampling methods inevitably carry the risk of introducing synthetic noise. Future work may therefore benefit from investigating more advanced techniques such as GAN-based data augmentation or adaptive sampling strategies to further improve detection performance.

Future research can extend this work in several directions. Expanding the dataset to include multiple exchanges and cross-market manipulations would enable broader generalization. Incorporating temporal deep learning architectures (e.g., LSTM, Temporal Convolutional Networks) or graph-based methods could better capture the complex, dynamic structure of trading networks. Moreover, integrating these approaches into real-time monitoring systems for exchanges, regulatory bodies, and financial institutions offers a promising avenue for practical implementation. Finally, from a managerial and governance perspective, future studies could explore how such detection models can be embedded within enterprise-level internal control frameworks to create AI-assisted early warning systems that strengthen market supervision, reduce operational risk, and enhance trust in digital asset ecosystems.